\documentclass{SCIS2023}

\usepackage{threeparttable}
\usepackage{tabularx}
\usepackage{booktabs}
\usepackage{verbatim}
\usepackage{amsopn}
\usepackage{amsmath}
\usepackage{multirow}
\usepackage{colortbl}
\usepackage{graphicx}
\usepackage{wrapfig}

\usepackage{xspace}

\makeatletter
\DeclareRobustCommand\onedot{\futurelet\@let@token\@onedot}
\def\@onedot{\ifx\@let@token.\else.\null\fi\xspace}

\def\eg{\emph{e.g}\onedot} 
\def\ie{\emph{i.e}\onedot} 
 
\def\etc{\emph{etc}\onedot} \def\vs{\emph{vs}\onedot}
 
\def\etal{\emph{et al}\onedot}
\makeatother
\usepackage{epstopdf}

\begin{document}
\ArticleType{RESEARCH PAPER}
\Year{}
\Month{}
\Vol{}
\No{}
\DOI{}
\ArtNo{}
\ReceiveDate{}
\ReviseDate{}
\AcceptDate{}
\OnlineDate{}

\title{ViGT: Proposal-free Video Grounding with Learnable Token in Transformer}{ViGT: Proposal-free Video Grounding with Learnable Token in Transformer}

\author[1,2]{Kun LI}{}
\author[1,2,3,4]{Dan GUO}{{guodan@hfut.edu.cn}}
\author[1,2,3,4]{Meng WANG}{eric.mengwang@gmail.com}

\AuthorMark{Li K}

\AuthorCitation{Li K, Guo D, Wang M}


\address[1]{School of Computer Science and Information Engineering, Hefei University of Technology, Hefei {\rm 230601}, China}
\address[2]{Key Laboratory of Knowledge Engineering with Big Data, Ministry of Education, Hefei {\rm 230601}, China}
\address[3]{Intelligent Interconnected Systems Laboratory of Anhui Province, Hefei {\rm 230601}, China} 
\address[4]{Institute of Artificial Intelligence, Hefei Comprehensive National Science Center, Hefei {\rm 230088}, China}

\abstract{The video grounding (VG) task aims to locate the queried action or event in an untrimmed video based on rich linguistic descriptions. Existing proposal-free methods are trapped in complex interaction between video and query, overemphasizing cross-modal feature fusion and feature correlation for VG. 
In this paper, we propose a novel boundary regression paradigm that performs regression token learning in a transformer. Particularly, we present a simple but effective proposal-free framework, namely \emph{\textbf{Vi}}deo \emph{\textbf{G}}rounding \emph{\textbf{T}}ransformer (\emph{\textbf{ViGT}}), which predicts the temporal boundary using a learnable regression token rather than multi-modal or cross-modal features. 
In \emph{\textbf{ViGT}}, the benefits of a learnable token are manifested as follows. (1) The token is unrelated to the video or the query and avoids data bias toward the original video and query. (2) The token simultaneously performs global context aggregation from video and query features.
First, we employed a sharing feature encoder to project both video and query into a joint feature space before performing cross-modal co-attention (\ie, video-to-query attention and query-to-video attention) to highlight discriminative features in each modality. Furthermore, we concatenated a learnable regression token [REG] with the video and query features as the input of a vision-language transformer. Finally, we utilized the token [REG] to predict the target moment and visual features to constrain the foreground and background probabilities at each timestamp. The proposed \emph{\textbf{ViGT}} performed well on three public datasets: ANet Captions, TACoS and YouCookII. Extensive ablation studies and qualitative analysis further validated the interpretability of \emph{\textbf{ViGT}}.
}

\keywords{video grounding, temporal sentence grounding, boundary regression, token learning, proposal-free.}

\maketitle

\section{Introduction}
Recently, Video grounding (VG), which is a challenging task in the computer vision community~\cite{chen2022fast,wang2022survey}, has received increasing attention~\cite{gao2017tall,yuan2019find,zhang2020vslnet,zhang2020learning,li2021proposal}. 
The associated tasks include action classification~\cite{wang2016temporal}, action localization~\cite{shou2016temporal,buch2017sst}, \etc. Unlike these action-relevant tasks, which recognize or locate the action in videos automatically or with the predefined action labels, the VG task in our work requires first understanding a linguistic query sentence and then locating the queried visual content in the video. 
Therefore, the VG task has also attracted considerable interest due to its importance in the field of vision-language understanding, such as video captioning~\cite{chen2020learning}, video question answering~\cite{li2022equivariant}, and cross-modal understanding~\cite{ji2022heterogeneous,guo2018hierarchical,qu2017novel,guo2019dense}. 
The conventional paradigm of VG~\cite{gao2017tall,yuan2019find,mun2020local} encodes video and query features, employs different methods of multi-modal interaction or multi-modal fusion to obtain cross-modal features, and finally predicts the target video segment. 
As illustrated in Figure~\ref{fig:abs}, existing methods can be classified into \textit{proposal-based} methods and \textit{proposal-free} methods. 

\begin{small}
\begin{figure}[t]
\centering
\includegraphics[width=0.6\linewidth]{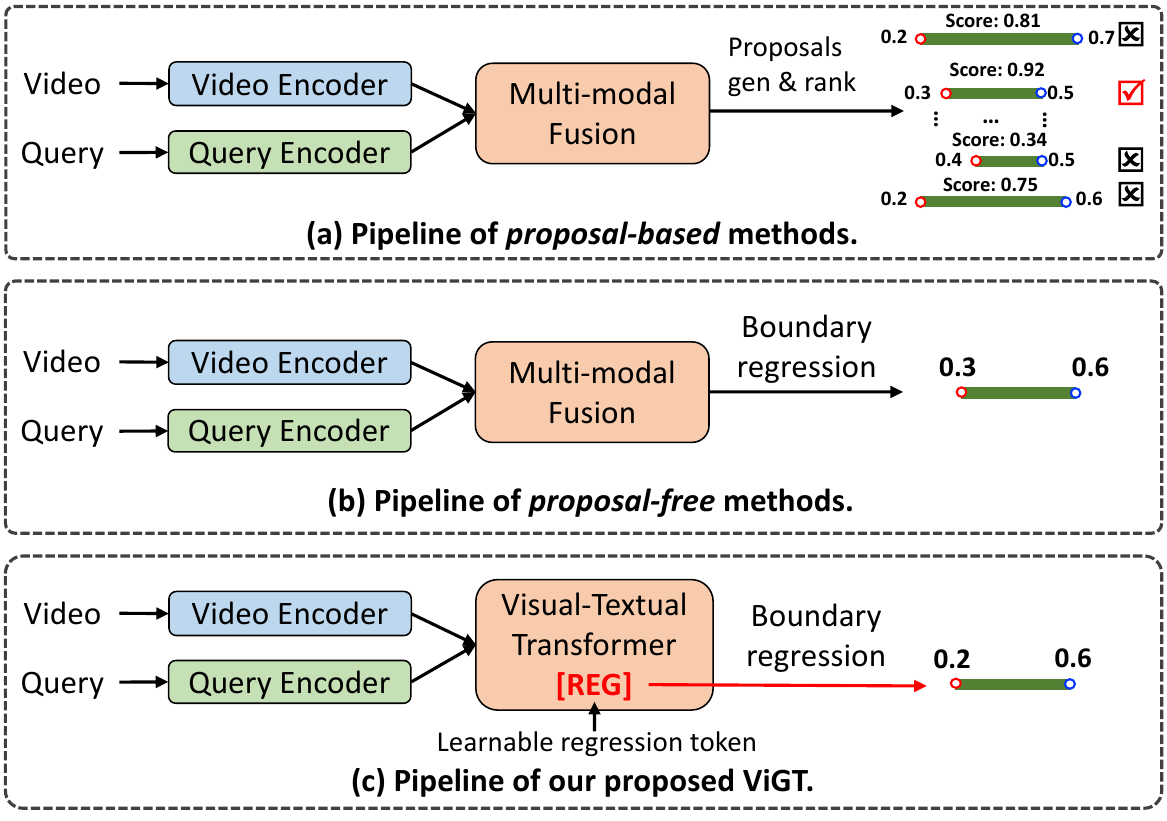}
\caption{Pipeline of the video grounding task. Previous works can be divided into two pipelines: (a) proposal-based methods (\eg,~\protect\cite{gao2017tall,yuan2019semantic}) and (b) proposal-free methods (\eg,~\protect\cite{li2021proposal,mun2020local,yuan2019find}). In contrast to them, the (c) proposed \emph{\textbf{ViGT}} performs the boundary regression with a learnable token in transformer architecture. 
}
\label{fig:abs}
\end{figure}
\end{small}

In {\bf proposal-based methods}, numerous candidate proposals are generated to cover the target moment. The researchers then apply the confidence scores of these dense candidate proposals to choose the final proposals. In the early stage, VG was implemented using clip-query matching~\cite{gao2017tall,anne2017localizing,liu2018attentive,liu2018cross}. 
It is assumed that cropped video segments can cover the target segment by applying a multi-scale sliding window on the videos. Under this assumption, temporal Intersection over Union, tIoU~\cite{gao2017tall} is constantly calculated for binary action or event classification evaluation. 
However, this assumption is not always valid, particularly when the target segments vary over a wide range. 
To generate high quality proposals, Chen \etal \cite{chen2018temporally} and Wang \etal \cite{wang2020temporally} produced proposals with multiple times scales at each time by following action detection~\cite{buch2017sst}. Zhang \etal \cite{zhang2020learning} enumerated all the candidate segments using a 2D temporal map. Furthermore, Liu~\etal~\cite{liu2021context} proposed a novel bi-affine network to predict all possible candidates. 
\emph{\bf Proposal-free methods} directly predict the target moment without any candidate proposals, avoiding the heavy calculation of candidate proposal generation and ranking. Yuan \etal \cite{yuan2019find} and Mun \etal \cite{mun2020local} predicted the segment boundaries using query-to-video attention. 
Rodriguez~\etal~\cite{rodriguez2020proposal} and Zhang~\etal~\cite{zhang2020vslnet} predicted the probability scores of start and end at each timestamp and then chose the position with the highest ones as the target moment. Chen~\etal~\cite{chen2020learning} performed both sequence-wise and channel-wise modality interactions. Li~\etal~\cite{li2021proposal} realized the multi-scale 2D correlation map in a proposal-free manner.

Even though previous works have made considerable advances in VG, they have some weaknesses. Typically, \emph{proposal-based methods} are trapped in numerous candidate proposal generation and human-heuristic proposal ranking rules, making the model overfit to the distribution of training data. 
Additionally, the accuracy of proposal ranking is easily influenced by the candidate proposals and the post-processing efficiency. 
Current \emph{proposal-free methods} largely depend on cross-modal fusion or interaction methods, which are always solved in two modes: 1) first one-side intra-modal correlation and then fusion, or 2) mapping a query to a semantic vector and then correlating it with visual features. 
\emph{Proposal-free methods} gets rid of redundant candidate proposals to reduce the costly computational overhead; however, they are easily influenced by the fusion or correlation effect of video and query. 

\begin{figure}
\centering
\includegraphics[width=0.6\linewidth]{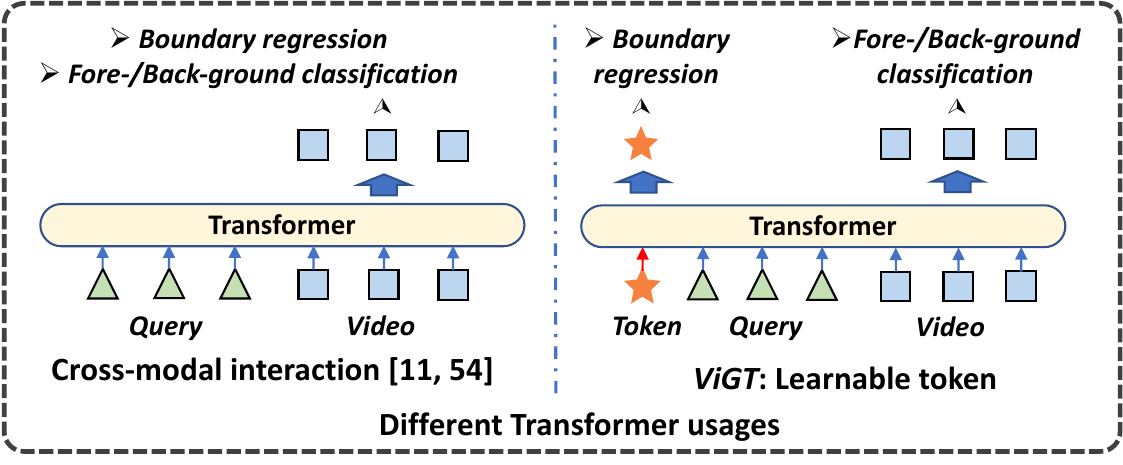}
\caption{Different transformer-based architectures for video grounding. In comparison to current works~\cite{chen2021end,zhang2021multi}, the proposed \emph{\textbf{ViGT}} has a different usage. We investigated a learnable token rather than multi-modal features for boundary regression. 
}
\label{fig:usage}
\end{figure}

In this work, we propose a simple but effective transformer-based framework for exploring a proposal-free regression using a learnable token. Our method is implemented in an end-to-end manner. We named the proposed method with a learnable Token in the transformer for VG as \emph{\textbf{ViGT}}. 
The pipeline of \emph{\textbf{ViGT}} is depicted in Figures~\ref{fig:abs} and~\ref{fig:usage}. 
Unlike current proposal-based methods, the proposed \emph{\textbf{ViGT}} no longer requires candidate proposals but predicts the target moment from a learnable token. 
Furthermore, the visual-linguistic correlation is addressed by imposing a learnable token on the relation learning of video and query rather than multi-modal or cross-modal features. 

The overview of \emph{\textbf{ViGT}} is depicted in Figure~\ref{fig:method}. First, we encode encode the video and query features using a video-language encoder. Next, the learnable regression token is concatenated with video and query features. It is then encoded by a video-language transformer. Finally, the regression token is used to predict the target moment. We attribute the success of the learnable token in our work to two aspects as follows: 
(1) The token is unrelated to the video or the query. It is randomly initialized and continually updated during model optimization. It prevents data bias toward the original video and query. (2) The token simultaneously interacts with all video and query features and retains informative cues in a global viewpoint of contextual aggregation.

The main contributions of our method are summarized as follows:
\begin{itemize}
\item We consider the concept of a learnable token for VG. We propose a simple but effective proposal-free transformer-based framework that implicitly captures the relationship between the video and query. 
\item The proposed method is implemented in a proposal-free manner, and it predicts the target moment from a learnable token rather than multi-modal or cross-modal features, as in previous works. 
\item Extensive experiments are conducted on three benchmark datasets and demonstrate the efficacy of the proposed method. Ablation studies and qualitative visualizations also confirm the contribution of each component.
\end{itemize}

\section{Related Work}\label{sec:related}
This section reviews the two VG pipelines: proposal-based and proposal-free methods. Our method falls under the category of proposal-free methods. Furthermore, we also review the related transformer-based work for video understanding because our method is a transformer-based network.

\subsection{Video Grounding}
The VG task evolved from temporal action detection, in which the action instance is located in the video, and its action category is identified. 
Later, the action detection task was extend to VG with a language sentence by the researchers. The VG task is also well known as temporal sentence grounding~\cite{liu2021context,yuan2019semantic}, video moment retrieval~\cite{zhang2019man,liu2023survey}, and temporal activity localization via language~\cite{yang2020survey}
, \etc. 
We grouped the existing works into proposal-based and proposal-free methods. 

\subsubsection{Proposal-based methods} To accurately locate the target action moment in videos, a simple solution is to generate a large number of potential proposals of activity instances and then select the best matching proposal as the final prediction. 
Early works~\cite{gao2017tall,anne2017localizing,liu2018attentive,liu2018cross} were always performed in a \emph{propose-and-rank} manner. They used a multi-scale sliding window to generate proposal candidates and then measured the semantic distance to select the best one. However, the sliding-window-based methods suffer from heavy calculations and huge memory costs. 
To address these issues, some works~\cite{xu2019multilevel,chen2018temporally} proposed to generate proposals that are directly conditioned on the sentence query. 
This proposal generation strategy was more adaptable than the sliding window strategy.
Particularly, Xu~\etal~\cite{xu2019multilevel} proposed a query-guided segment proposal network that leverages video and query correlation to generate temporal attention weights for proposal generation. 
Chen~\etal~\cite{chen2018temporally} proposed a semantic activity proposal network that generated proposals based on a temporally semantic correlation score between video and semantic entities in the query.  
However, the quality of generated proposals typically depended on the multi-modal interaction in the proposal generation module.
Motivated by the proposal generation methods SSD for object detection~\cite{liu2016ssd} and SST for action detection~\cite{buch2017sst}, several similar works were proposed for VG~\cite{chen2018temporally,wang2020temporally,yuan2019semantic}. 
Based on SST, TGN~\cite{chen2018temporally} extended a cross-modal interaction before proposal generation for VG. Inspired by the SSD model, SCDM~\cite{yuan2019semantic} utilized a stacked convolutional block to generate dense proposals through iterative interaction. Specifically, multi-scale proposals were generated by using the stacked temporal convolutional block, which can cover the varied-length target actions. 
Additionally, unlike the aforementioned methods, Zhang~\etal~\cite{zhang2020learning} proposed a 2D temporal map to enumerate all proposals, while Zeng~\etal~\cite{zeng2020dense} addressed the imbalance problem of temporal boundaries distribution of ground-truth through a dense regression head. In summary, these proposal-based methods have made considerable progress through efficient proposal generation. 
However, the proposal-based methods can not get rid of the post-processing of proposal ranking, which typically reduces grounding efficiency. 

\subsubsection{Proposal-free methods}
As opposed to proposal-based methods, proposal-free methods do not require the generation and ranking of candidate proposals but directly predict the start and end timestamps. Proposal-free approaches are more efficient and can be easily applied to videos of varying lengths. \emph{On the one hand}, some approaches predict the probability scores of starting and ending timestamps along the timeline, and choose the highest one as the target moment, such as TMLGA~\cite{rodriguez2020proposal}, VSLNet~\cite{zhang2020vslnet}, DORi~\cite{rodriguez2021dori}. 
\emph{On the other hand}, some approaches focus on moment regression, namely directly outputting the final moment's starting and ending times, such as ABLR~\cite{yuan2019find}, LGI~\cite{mun2020local}, CPNet~\cite{li2021proposal}, PMI~\cite{chen2020learning}, and HVTG~\cite{chen2020hierarchical}. 
For the score probability-based approaches, TMLGA~\cite{rodriguez2020proposal} and VSLNet~\cite{zhang2020vslnet} projected the query onto the video and then used multi-modal features to predict the probabilities of starting and ending at each timestamp. 
However, following the moment interval labels, the annotation value of the score curve is 1 only at the starting/ending positions; otherwise, 0. In other words, the supervision information provided along the timeline is limited and insufficient. This makes it difficult for the model to predict exact score probabilities of starting and ending along the timeline. 
Furthermore, visual reasoning was considered in this task. DORi~\cite{rodriguez2021dori} designed a language-conditioned spatial graph to model the fine-grained relationship between the visual elements of objects and subjects in the videos for visual reasoning. 
The object information indeed helps in building fine-grained video-text interaction modeling, but the fine-grained feature extraction at the object-level requires additional computation and memory.
In terms of regression approaches, LGI~\cite{mun2020local} developed a local-global video-text interaction network that performed the interactions between video and multiple phrases of query in a local-to-global manner for boundary regression. Nonetheless, the language query is divided into multiple phrases, which may limit the comprehension of the query.  
PMI~\cite{chen2020learning} presented a pairwise interaction network that models the modality interaction at the pair of sequence-level and channel-level semantics. 
HVTG~\cite{chen2020hierarchical} used a hierarchical graph composed of object-object and object-sentence subgraphs to model fine-grained visual-textual interactions for VG. Both the above DORi~\cite{rodriguez2021dori} and HVTG~\cite{chen2020hierarchical} took fine-grained object information into account. 
DORi~\cite{rodriguez2021dori} predicted the score probabilities while HVTG~\cite{chen2020hierarchical} implemented the moment regression. 
These methods heavily rely on object detection and have complex calculations, especially in complex scenes. 
By further observation, in current proposal-free methods, a core technique is the attention mechanism, \eg, cross-attention~\cite{rodriguez2020proposal,yuan2019find}, co-attention~\cite{zhang2020vslnet,li2021proposal}, self-attention~\cite{mun2020local}, transformer~\cite{zhang2021multi}, and spatial-temporal graph~\cite{rodriguez2021dori} that are all explored to address this task. The purpose of attention is to enhance visual or textual features through various interaction mechanisms. 
Furthermore, there are some new technique attempts, \eg, reinforcement learning-based method to design the reward policy for this tasks~\cite{he2019read,wang2019language}, contrastive learning mechanism with the structured causal model to learn more representative features~\cite{nan2021interventional}.

\subsection{Transformer for Video Understanding}
The booming technique transformer is rapidly developed in the field of vision and language, such as famous pre-training architectures DETR~\cite{carion2020end}, ViT~\cite{dosovitskiy2020image}, and VL-BERT~\cite{su2019vl}. 
These pre-training architectures are typically based on large-scale datasets, and have a large number of parameters and calculation computations. Typically, for visual and video understanding tasks, there are TransVG for visual grounding~\cite{deng2021transvg}, ViViT for action classification~\cite{arnab2021vivit} and DETR~\cite{carion2020end} for moment detection~\cite{lei2021detecting}, \etc. 
As for the VG task, Zhang~\etal~\cite{zhang2021multi} proposed a multi-stage aggregated transformer network (MATN), which utilized the architecture VL-BERT~\cite{su2019vl} to enhance cross-modal features, 
and Chen~\etal~\cite{chen2021end} proposed a multi-modal learning framework named DRFT that leveraged a co-attentional transformer and contrastive learning to learn more informative cross-modal features. 
In the both two studies, the methodological contributions were made to enhance query-video interaction. 
In other words, current transformer-based research is primarily concerned with how to improve the cross-modal representation ability. 
The learning of cross-modal correlation improves performance by enhancing visual and textual features. 
To summarize, both MATN and DRFT predict the target segment using cross-modal features. From the pipeline perspective, MATN~\cite{zhang2021multi} belongs to the proposal-based method, whereas DRFT~\cite{chen2021end} belongs to the proposal-free method. 
In contrast to these works, in this work, we propose an end-to-end proposal-free boundary regression paradigm using \emph{a learnable regression token} for VG (described in Section~\ref{src:method}). The main idea is that we use a learnable token in transformer architecture to detect and retain the relational contexts between the video and query. When compared to multi-modal features for boundary regression in current attention-based and transformer-based modes, the learnable token regression paradigm is a simple innovative attempt at VG. 

\section{Our Approach}
\label{src:method}
In this work, we propose a proposal-free video grounding framework through a learnable regression token in an end-to-end manner. We deem the video grounding task as a boundary regression problem. A model (denoted as $\mathcal{F}$) is required to locate the starting and ending timestamps of the target action or event segment described by language query $Q$ in video $V$:
\begin{equation}
<\hat{t}_s, \hat{t}_e > =\mathcal{F}(V, Q, \Theta),
\end{equation}
where $\hat{t}_s$ and $\hat{t}_e$ denote the starting and ending timestamps respectively, and 
$\Theta$ is the parameter collection of model $\mathcal{F}$. 

\begin{figure*}[t]
\centering
\includegraphics[width=1.0\columnwidth]{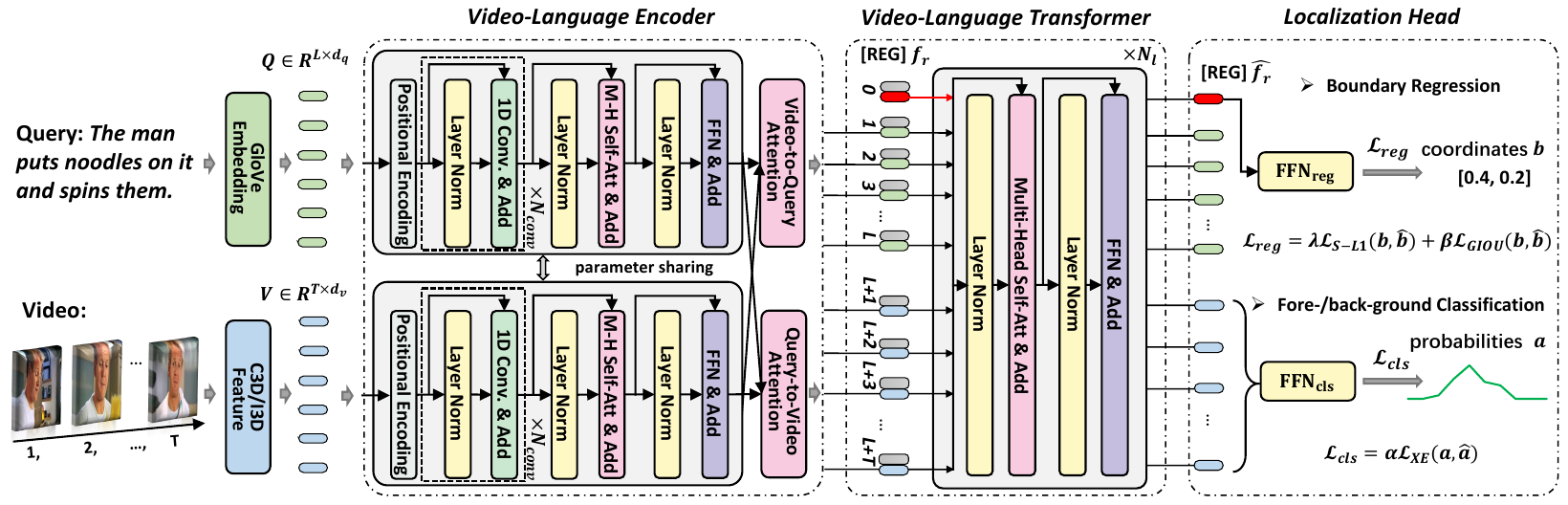}
\caption{Overall framework of \textbf{\emph{ViGT}} for video grounding. ``M-H Self-Att'' in video-language encoder denotes Multi-Head Self-Attention operator. First, we use a sharing multi-modal feature encoder to map video and query features into a joint feature space. Subsequently, we investigate a co-attention mechanism (\ie, query-to-video and video-to-query attentions) to correlate these two features. Next, the correlated query and video features are concatenated with a [REG] token and fed to a Video-Language Transformer. Finally, merely the [REG] token is used to predict the target boundary (defined as a boundary head FFN$_{reg}$). Simultaneously, the video features are used to constrain foreground or background along the timeline of video (defined as classification head FFN$_{cls}$). In other words, both the two heads are used during training, while we just use the boundary regression head FFN$_{reg}$ for prediction.}
\label{fig:method}
\end{figure*}

The overview of our proposal-free video grounding with learnable token method 
(\textbf{\emph{ViGT}}) is shown in Figure~\ref{fig:method}. The \textbf{\emph{ViGT}} is inspired by the famous transformer architecture. For this task, we elaborate the proposed method as below: 1) we employ a sharing feature encoder to map both video and query sequence into a joint feature space, 2) then we perform cross-modal co-attention to highlight the discriminative features in each modality, 3) more importantly, we concatenate a learnable regression token [REG] with both query and video features as the input of vision-language transformer, and 4) finally, we utilize the regression token [REG] to predict the boundary. Besides, we further use the video features to constrain the probabilities of foreground and background along the timeline of video, where the foreground denotes the occurrence of the queried action or event in the video. 

\subsection{Video-Language Encoder}
As a preliminary operation, we extract original query features $Q\in\mathbb{R}^{L\times d_q}$ by GloVe embedding~\cite{pennington2014glove} and original video features $V\in\mathbb{R}^{T\times d_v}$ by using the pre-trained C3D~\cite{tran2015learning} or I3D~\cite{carreira2017quo}. Then, we perform a video-language encoder to further encode the intra- and inter-modal relationships in and between the video and query.

\subsubsection{Feature Encoding}
In this study, we use the transformer-encoder in QANet~\cite{yu2018qanet} as a \textbf{Feature Encoder} ({\bf FE}) layer to embed both the query and video sequences. The encoder consists of a stacked convolutional block for local context learning, multi-head self-attention (${\bf MSA}$) for long-dependence learning, positional encoding (${\bf PE}$)~\cite{vaswani2017attention} for position modeling, and a feedforward layer (${\bf FFN}$) for feature projection, respectively. Besides, layer norm (${\bf LN}$) is also applied. Let the input feature sequence as $\bm{X}\!=\!\{x_1, x_2, x_3, \ldots, x_N\}\in\mathbb{R}^{N\times d}$, the feature encoder is formulated as follow: 
\begin{equation}
\begin{aligned}
\bm{\hat{X}} &= {\bf Feature\ Encoder}(\bm{X}) \Leftrightarrow\\
&\left\{\begin{array}{l}
	\bm{X^{'}} = {\bf PE}(\bm{X}) + \bm{X} ;\\
	\bm{\bar{X}} = {\bf Conv1D}({\bf LN}( \bm{X^{'}}), K_{1D}, N_{conv}) +  \bm{X^{'}};\\ 
	\bm{\tilde{X}} = {\bf MSA}(
	{\bf LN}(\bm{\bar{X}}))\!+\!\bm{\bar{X}} ;\\
	\bm{\hat{X}} = {\bf FFN}({\bf LN}(\bm{\tilde{X}})) + \bm{\tilde{X}},
\end{array}\right.
\end{aligned}
\end{equation}
where ${\bf Conv1D}(\cdot, K_{1D}, N_{conv})$ implements one-dimensional convolution with kernel size of $K_{1D}$ and layer number of $N_{conv}$. 

Given the original features of video $\bm{V}$=$\{v_1$, $v_2$, $v_3$, $\ldots$, $v_T\}$ $\in\mathbb{R}^{T\times d_v}$ and query $\bm{Q}$=$\{q_1, q_2, q_3,$$ \ldots, q_L\}$$\in\mathbb{R}^{L\times d_q}$, we project them into the same dimension $d$:
\begin{equation}
\left\{\begin{array}{l}
\hat{\bm{Q}} = {\bf Feature\ Encoder} (\bm{QW_q}) \in \mathbb{R}^{L\times d};\\
\hat{\bm{V}} = {\bf Feature\ Encoder} (\bm{VW_v}) \in \mathbb{R}^{T\times d},
\end{array}\right.
\label{equ:FE}
\end{equation}
where $\bm{W_q} \in \mathbb{R}^{d_q\times d}$ and $\bm{W_v}\in \mathbb{R}^{d_v\times d}$ are two learnable parameters for linear projection. Please note that the parameters of this \textbf{Feature Encoder} ({\bf FE}) layer are shared between video and query. Up to now, we obtain new encoded features  $\hat{\bm{V}}\in\mathbb{R}^{T\times d}$ and $\hat{\bm{Q}}\in\mathbb{R}^{L\times d}$.

\subsubsection{Cross-Modal Co-Attention}
The above feature encoder learns intra-modal relationship in each modality. Here, we use a {\bf Cross-Modal Co-Attention} ({\bf CMCA}) to capture the inter-modal correlation between video and query. The {\bf CMCA} is implemented with a vanilla transformer~\cite{vaswani2017attention}. 
Taking video-to-query attention as an example, we target to discover responsive visual features $\bm{\hat{V}}$ under the guidance of query $\bm{\hat{Q}}$, and aggregate them onto the textual dimension. Thus, $\bm{Q^*}$ is a new textual feature with video aggregation formulated as below:
\begin{equation}
\begin{aligned}
\bm{Q^*} &= {\bf CMCA}(\bm{\hat{Q}}, \bm{\hat{V}})\Leftrightarrow \\
&\left\{\begin{array}{l}
	\bm{Q^{''}} = {\bf LN}\! \big({\bf MSA}(\!\mathcal{Q}\!=\bm{\hat{Q}}; \mathcal{K}\!,\!\mathcal{V}\!=\!\bm{\hat{V}})\!+\!\bm{\hat{Q}}\big) ;\\
	\bm{Q^*} = {\bf LN}\big({\bf FFN}(\bm{Q^{''}}) + \bm{Q^{''}}\big),
\end{array}\right.
\end{aligned}
\label{equ:CMCA}
\end{equation}
where $\mathcal{Q}$, $\mathcal{K}$, and $\mathcal{V}$ are three factors in the \textbf{MSA} operation.

Similarly, for query-to-video attention, 
taking video $\bm{\hat{V}}$ as a query, we discover the responsive query features in $\bm{\hat{Q}}$ and aggregate them onto the visual dimension. Thus, a new video feature with query aggregation is calculated as follows: 
\begin{equation}
\begin{aligned}
\bm{V^*}&={\bf CMCA}(\bm{\hat{V}} , \bm{\hat{Q}})\Leftrightarrow \\
&\left\{\begin{array}{l}
	\bm{V^{''}} = {\bf LN}\! \big({\bf MSA}(\!\mathcal{Q}\!=\bm{\hat{V}}; \mathcal{K}\!,\!\mathcal{V}\!=\!\bm{\hat{Q}})\!+\!\bm{\hat{V}}\big) ;\\
	\bm{V^*} = {\bf LN}\big({\bf FFN}(\bm{V^{''}}) + \bm{V^{''}}\big).
\end{array}\right.
\end{aligned}
\label{equ:CMCA_q}
\end{equation}

\subsection{Video-Language Transformer}
After the above-mentioned feature encoding, the enhanced feature representation is more discriminative. Here, we add a learnable token [REG] and apply a video-language transformer to model the relationships among the token, query and video features. Specifically, the video-language transformer consists of $N_{l}$ transformer encoder blocks with multi-head self-attention (\textbf{MSA}), feedforward layer (\textbf{FFN}), layernorm (\textbf{LN}), and residual connections. Let the regression token [REG] be $f_r\in\mathbb{R}^{d}$, we concatenate it with the video feature $\bm{V^*}=[v_{1}^{*}, \cdots, v_{T}^{*} ]$ and query feature $\bm{Q^*}=[q_{1}^{*}, \cdots, q_{L}^{*}]$. The transformer block is formulated as follows:
\begin{equation}
\left\{\begin{array}{l} 
{\bf z}_0 = [f_r; q_{1}^{*}, \cdots, q_{L}^{*}; v_{1}^{*}, \cdots, v_{T}^{*} ]\!+\!f_{pos};\\
{\bf z'}_n = {\bf MSA}({\bf LN}({\bf z}_{n-1})) + {\bf z}_{n-1}; \\
{\bf z}_n = {\bf FFN}({\bf LN}({\bf z'}_n )) + {\bf z'}_n, 
\end{array}\right.
\end{equation}
where $n \in [1, N_{l}]$, $f_{pos}\!\in\! \mathbb{R}^{(L+T+1)\times d}$ denotes the initialized position embedding \cite{vaswani2017attention}, and we add it to the input feature sequence for temporal modeling. The regression token [REG] is randomly initialized and optimized with the full model.

\subsection{Localization Head}
In this work, we consider two types of localization head for video grounding, \ie, a boundary regression head and a fore-/back-ground classification head. For {\bf boundary regression}, we define a feedforward unit as ${\bf FFN_{reg}}$, which consists of three fully-connected layers and a \emph{sigmoid} activation. 
Through the regression head ${\bf FFN_{reg}}$, we obtain the predicted target moment. 
Specifically, we input the token [REG] $\hat f_r$ into ${\bf FFN_{reg}}$, 
where $\hat f_r\in \mathbb{R}^{d}$ is the output of video-language transformer as shown in Figure~\ref{fig:method}.
The predicted target moment $b$ is calculated by:
\begin{equation}
b = {\bf FFN_{reg}} (\hat f_r),
\label{equ:reg_head}
\end{equation}
where $b\in [0, 1]^2$ denotes the normalized center and width coordinates of the target moment. 

To achieve the semantic alignment between video and query, we design an additional head, \ie, {\bf fore-/back-ground classification head} denoted as ${\bf FFN_{cls}}$. 
We assign 1 to foreground (\ie, target segment) and 0 to background, and then predict a confidence score $a\in \mathbb{R}^{T}$ along the timeline of video. In other words, $a$ is used to judge each visual feature whether belong to foreground or not (background). Here, we design a feedforward unit consisting of a layer fully-connected layer and a \emph{sigmoid} activation as ${\bf FFN_{cls}}$. 
\begin{equation}
a = {\bf FFN_{cls}} (\bm{\hat{V}^*}), 
\end{equation}
where $\bm{\hat{V}^*} \in \mathbb{R}^{T\times d}$ denotes the output visual features from the video-language transformer as shown in Figure~\ref{fig:method}.

\subsection{Training}
According to the localization head, we design a multi-task loss. For the \textbf{boundary regression head}, we set two objective terms. The first term is \emph{Smooth-l$_1$} loss, which is widely applied to estimate the target moment in previous works~\cite{yuan2019find,li2021proposal}. The second term is \emph{Generalized IoU} (GIoU) loss originated from object detection~\cite{rezatofighi2019generalized}. 
We apply it in this field to supervise the target boundary. The total regression loss $\mathcal{L}_{reg}$ is formulated as follows:
\begin{equation}
\mathcal{L}_{reg} = \lambda \mathcal{L}_{s\!-\!l_1} (\Phi(b), \Phi(\hat{b})) +\beta \mathcal{L}_{giou}(b, \hat{b}),
\label{loss:reg}
\end{equation}
where $\lambda$ and $\beta$ are trade-off hyper-parameters, $\hat{b}\in [0, 1]^2$ is the normalized center and width coordinates of target moment in ground-truth, and function $\Phi(\cdot)$ converts the target moment from the format $<$center, width$>$ to format $<$start, end$>$. For the GIoU term, we calculate it with 1D temporal boundary $<$center, width$>$ by replacing 2D bounding box $<$center\_x, center\_y, width, height$>$ in the IoU calculation of object detection~\cite{rezatofighi2019generalized}. 

In addition, we define a classification loss $\mathcal{L}_{cls}$ to supervise the confidence score $a$ in the {\bf fore-/back-ground classification head}. The loss $\mathcal{L}_{cls}$ is implemented as follows:
\begin{equation}
\mathcal{L}_{cls} = \alpha \mathcal{L}_{XE}(a, \hat{a}),
\label{loss:cls}
\end{equation}
\begin{equation}
\mathcal{L}_{XE} (a, \hat{a}) = \frac{1}{N}\sum_i - [\hat{a}\cdot \log (a) + (1-\hat{a})\cdot \log(1-a)],
\end{equation}
where $\alpha$ is a balance hyper-parameter, $a$ denotes the predicted probability of each visual feature belong to foreground, and $\hat{a}$ is the ground-truth. 
To summarize, the multi-task loss is formulated as in Eq.~\ref{equ:loss}. Please note that with the two localization heads, $\mathcal L_{reg}$ and $\mathcal L_{cls}$ are both used during training. For target boundary prediction, we only use the boundary regression head for prediction.
\begin{equation}
\mathcal{L} = \mathcal{L}_{reg} + \mathcal{L}_{cls}.
\label{equ:loss}
\end{equation}

\begin{table}[t]
\centering
\footnotesize
\tabcolsep 9pt
\caption{Performance comparison on the ANet-Captions dataset with C3D features. ``Core Tech" records the methodological technique. ``Transformer with MLM" denotes using Masked Language Modeling (MLM) in the transformer training. DRFT$^\dagger$ denotes the DRFT model trained with rich features of RGB, depth and optical flow data.}
\begin{tabular}{c|c|c|c|ccc|c}
\toprule
&   Method    &   Venue    &       Core Tech    & IoU@0.3$\uparrow$   & IoU@0.5$\uparrow$ & IoU@0.7$\uparrow$ & mIoU$\uparrow$ \\ \hline
&     MCN~\cite{anne2017localizing}     &  ICCV'17   &     Sliding window     & 39.35  &    21.36       &       6.43        &     15.83      \\
&    CTRL~\cite{gao2017tall}     &  ICCV'17   &     Sliding window     & 47.43   &    29.01       &       10.34       &     20.54      \\
&     TGN~\cite{chen2018temporally}    &  SIGIR'18  &      Interaction      & 45.51  &     28.47       &        --         &       --       \\
&    SCDM~\cite{yuan2019semantic}    &  NIPS'19   &      Interaction      & 54.80   &    36.75       &       19.86       &       --       \\
&    CMIN~\cite{zhang2019cross}     &  SIGIR'20  &       Attention        & 63.61   &   43.40       &       23.88       &       --       \\
&     CBP~\cite{wang2020temporally}    &  AAAI'20   &      Interaction      &  54.30  &   35.76       &       17.80       &     36.85      \\
&   2D-TAN~\cite{zhang2020learning}  &  AAAI'20   &    2D Temporal Map     &  59.45  &   44.05       &       27.38       &       --       \\
&     DRN~\cite{zeng2020dense}    &  CVPR'20   &    Dense regression    & -- &   45.45       &       24.36       &       --       \\
& MATN~\cite{zhang2021multi} &  CVPR'21   &      Transformer       &  --   &  46.26       &       \underline{28.82}       &       --       \\
&    MATN$^\ddagger$~\cite{zhang2021multi}     &  CVPR'21   &  Transformer with MLM  &  --  &   \underline{48.02}       &   \underline{31.78}   &       --     \\  
\multirow{-8}{*}{
	\rotatebox{90}{\emph{Proposal-based methods}}} &    CBLN~\cite{liu2021context}     &  CVPR'21   &   Multi-interaction &     \underline{66.34}  &   \underline{48.12}       &  \underline{27.60}       &       --       \\ 
& MI-PUL~\cite{ding2022exploring} & TIP'22 & 2D Temporal Map & 60.15 & 46.35 & \underline{28.13} & --\\ 
& MGPN~\cite{sun2022you} & SIGIR'22 &  2D Temporal Map & -- & \underline{47.92} & \underline{30.47} & -- \\
& HCLNet~\cite{zhang2022video} & ACM MM'22 &  Contrastive learning & 63.05 & 45.82 & 26.79 & 44.89 \\
\bottomrule  
&   Method    &   Venue    &        Core Tech   & IoU@0.3$\uparrow$      & IoU@0.5$\uparrow$ & IoU@0.7$\uparrow$ & mIoU$\uparrow$ \\ \hline
&    ABLR~\cite{yuan2019find}  &  AAAI'19   &       Attention        & 55.67     &   36.79       &        --         &     36.99      \\
&  DEBUG~\cite{lu2019debug} & EMNLP'19   &  Attention & 55.91 & 39.72 & -- &  39.51 \\
&   TripNet~\cite{hahn2019tripping}  &  BMVC'20   & Reinforcement Learning &  48.42 &      32.19       &       13.93       &       --       \\
&    TMLGA~\cite{rodriguez2020proposal}    &  WACV'20   &       Attention        & --  &      33.04       &       19.26       &     37.78      \\
&     LGI~\cite{mun2020local}     &  CVPR'20   &       Attention        & 58.52  &      41.51       &       23.07       &     41.13      \\
&     PMI~\cite{chen2020learning}     &  ECCV'20   &      Interaction      & 59.69  &      39.37       &       19.27       &       --       \\
&    HVTG~\cite{chen2020hierarchical}     &  ECCV'20   &  Visual-textual Graph  & 57.60  &      40.15       &       18.27       &       --       \\
&   VSLNet~\cite{zhang2020vslnet}    &   ACL'20   &       Attention        &  63.16  &        43.22       &       26.16       &     43.19      \\
&    DORi~\cite{rodriguez2021dori}     &  WACV'21   & Spatial-temporal Graph &  57.89 &        41.49       &       26.41       &     42.78      \\
&    BPNet~\cite{xiao2021boundary}    &  AAAI'21   &    2D Temporal Map     & 58.98 &      42.07       &       24.69       &     42.11      \\
&    CPNet~\cite{li2021proposal}    &  AAAI'21   &       Attention        & --   &     40.56       &       21.63       &     40.65      \\
& DRFT~\cite{chen2021end}        & NeurIPS'21 &      Transformer       & 60.25   &     42.37       &       25.23       &     43.18   \\
& DRFT$^\dagger$~\cite{chen2021end}  & NeurIPS'21 &      Transformer    &  62.91   &       45.72       &       \textbf{27.79}       &     \textbf{45.86} \\ 		
& IVG~\cite{nan2021interventional}  & CVPR'21 & Contrastive Learning  & 63.22 & 43.84 & 27.10 & 44.21 \\ \cline{2-8}
\multirow{-13}{*}{\rotatebox{90}{\emph{Proposal-free methods}}}    &    Ours     &     --     &      Transformer    &   \textbf{64.59}  &  \textbf{46.71 }  &  26.90   & 45.71 \\ \bottomrule
\end{tabular}
\label{tab:anet_results}
\end{table}
\section{Experiments}\label{sec:exp}
In this section, we make a comparison with previous works. Ablation studies are conducted to investigate the effectiveness of our method. 
We also show qualitative results to demonstrate the interpretability of our method. 

\subsection{Experimental setup} \label{sec:exp_setup}
\textbf{Datasets and Evaluation.} 
We experiment on three public benchmark datasets, and introduce datasets in detail as follows.

\begin{itemize}
\item \textbf{ActivityNet Captions (ANet-Captions)}~\cite{krishna2017dense} consists of 20K videos that are collected from YouTube. The average duration of video is about 2 minutes, and each video are annotated with 3.65 queries averagely. 
Following~\cite{zhang2020vslnet}, the ``train'' set is used for training, ``val\_1'' set for validation, and ``val\_2'' set for test. Specifically, the dataset is split into the train, validation and test sets of 37,421, 17,505, and 17,031 query-clip pairs.
We notice that this dataset is released for the dense video captioning task~\cite{krishna2017dense,wang2021end} including numerous segment-captioning pairs, and then is extended for video grounding. 

\item \textbf{TACoS}~\cite{gao2017tall} consists of more challenging  cooking videos. There are 127 average 4.79 minutes-long videos and each video has around 178 queries. Compared with the {ANet-Captions} dataset, the {TACoS} has denser queries on each video. This dataset is split int 10,146, 4,589, and 4,083 query-clip pairs for train, valid, and test sets, respectively.

\item \textbf{YouCookII}~\sloppy\cite{zhou2018towards} consists of 2,000 untrimmed long videos that are collected from YouTube. Videos in this dataset are collected in the cooking scenarios as in the TACoS dataset too. The average duration of video is 5.26 minutes and each video has 7.73 queries on average. 
\end{itemize}

\textbf{Evaluation Metrics.} Following the protocol~\cite{gao2017tall,mun2020local}, we adopt two metrics \textbf{``R@n, IoU@m''} to represent the ratio of queried moments having IoU (Interaction-Over-Union) larger than threshold $m$ in the top-$n$ predicted ones:
\begin{equation}
{\rm R@n, IoU@m}= \frac{1}{N_q}\sum_{i=1}^{N_q}r(n,m,q_i),
\end{equation}
where $N_q$ denotes the query number and $q_i$ denotes the $i$-th query. The rule is that $r(n,m,q_i)$=1 if the IoU value of $q_i$ with ground-truth is larger than $m$, otherwise $r(n,m,q_i)$=0.
\textbf{``mIoU''} represents the average IoU for all the queries. Since our \textbf{\textit{ViGT}} is a proposal-free method, we report the experimental results at $n$=1, where $m\in \{0.5, 0.7\}$ on the ActivityNet Captions and YouCookII datasets and $m \in \{0.3, 0.5\}$~\cite{zhang2020vslnet} on the TACoS dataset. \textbf{``R@n, IoU@m''} is abbreviated to \textbf{``IoU@m''} in experimental results. 

\begin{table*}[t]
\centering
\footnotesize
\tabcolsep 9pt
\caption{Performance comparison on the TACoS dataset with C3D features. $^{\star}$ denotes using I3D features provided by VSLNet~\cite{zhang2020vslnet}, and $^{\diamond}$ denotes using C3D features provided by 2D-TAN~\cite{zhang2020learning}. 
}
\begin{tabular}{c|c|c|c|ccc|c}
\toprule
&       Method        &  Venue   &       Core Tech   & IoU@0.1$\uparrow$     & IoU@0.3$\uparrow$ & IoU@0.5$\uparrow$ & mIoU$\uparrow$ \\ \hline
&         MCN~\cite{anne2017localizing}         & ICCV'17  &     Sliding window   &  -- &        --         &       5.58        &       --       \\
&        CTRL~\cite{gao2017tall}         & ICCV'17  &     Sliding window    & 24.32 &       18.32       &       13.30       &       --       \\
&         TGN~\cite{chen2018temporally}         & SIGIR'18 &      Interaction      &  41.87  &   21.77       &       18.90       &       --       \\
&        SCDM~\cite{yuan2019semantic}         & NIPS'19  &      Interaction      & --  &    26.11       &        --         &       --       \\
&        CMIN~\cite{zhang2019cross}         & SIGIR'20 &       Attention        &  32.48 &    24.64       &       18.05       &       --       \\
&         CBP~\cite{wang2020temporally}         & AAAI'20  &      Interaction      & --   &   27.31       &       24.79       &     21.59      \\
& 2D-TAN$^{\diamond}$~\cite{zhang2020learning} & AAAI'20  &    2D Temporal Map     &  47.59 &      37.29       &       25.23       &       --       \\
&         DRN~\cite{zeng2020dense}         & CVPR'20  &    Dense regression    &  --  &     --         &       23.17       &       --       \\
\multirow{-5}{*}{\rotatebox{90}{\emph{Proposal-based methods}}} &     MATN~\cite{zhang2021multi}     & CVPR'21  &      Transformer       &  -- &     45.64       &       34.79       &       --       \\
&        MATN$^\ddagger$~\cite{zhang2021multi}          & CVPR'21  &  Transformer with MLM  & --  &  \underline{48.79}   &   \underline{37.57}   &       --       \\
&        CBLN~\cite{liu2021context}         & CVPR'21  &   Multi-interaction   & 49.16    &   38.98       &       27.65       &       --       \\ 
& MI-PUL~\cite{ding2022exploring} & TIP'22 & 2D Temporal Map & 52.05 & 41.03 & 31.47 & --\\ 
& MGPN~\cite{sun2022you} & SIGIR'22 &  2D Temporal Map & -- & \underline{48.81} & \underline{36.74} & --\\
& HCLNet~\cite{zhang2022video} & ACM MM'22 & Contrastive learning & \underline{62.03} & \underline{50.04}& \underline{37.89} & \underline{34.80} \\
\bottomrule
&       Method        &  Venue   &       Core Tech       & IoU@0.1$\uparrow$  & IoU@0.3$\uparrow$ & IoU@0.5$\uparrow$ & mIoU$\uparrow$ \\ \hline
&        ABLR~\cite{yuan2019find}         & AAAI'19  &       Attention        &   -- &    19.50       &        --         &     13.40      \\
&       TripNet~\cite{hahn2019tripping}       & CVPRW'19 & Reinforcement Learning & --  &     23.95       &       19.17       &       --       \\
&        SM-RL~\cite{wang2019language}       & CVPR'19  & Reinforcement Learning & 26.51   &   20.25       &       15.95       &       --       \\
&        DEBUG~\cite{lu2019debug} & EMNLP'19   &  Attention & 41.15 & 23.45 & -- &  16.03 \\
&  VSLNet$^{\star}$~\cite{zhang2020vslnet}   &  ACL'20  &       Attention        & --    &  29.61       &       24.27       &     24.11      \\
&        DORi~\cite{rodriguez2021dori}         & WACV'21  & Spatial-temporal Graph & --   &   28.69       &       24.91       &     26.42      \\
&        BPNet~\cite{xiao2021boundary}        & AAAI'21  &    2D Temporal Map     &  --   &  25.96       &       20.96       &     19.53      \\
& CPNet$^{\diamond}$~\cite{li2021proposal}  & AAAI'21  &       Attention        & --   &   42.61       &       28.29       &     28.69      \\ 
& IVG~\cite{nan2021interventional}  & CVPR'21 & Contrastive Learning &   49.36 & 38.84  & 29.07 & -- \\  
\cline{2-8}
&   Ours$^{\star}$    &    --  &       Transformer    &  \textbf{57.56}  &  \textbf{45.34}   &  \textbf{31.77}   & \textbf{30.82} \\
\multirow{-10}{*}{\rotatebox{90}{\emph{Proposal-free methods}}}  &  Ours$^{\diamond}$  &    --    &      Transformer    &  \textbf{55.59} &  \textbf{45.99}   &  \textbf{32.32}   & \textbf{30.97} \\ \bottomrule
\end{tabular}

\label{tab:tac_results}
\end{table*}

\textbf{Implementation Details.}  
We conduct the same experiment setting on above three datasets. We use a pre-trained 3D network (\ie, C3D \cite{tran2015learning} or I3D~\cite{carreira2017quo}) to extract original video features. The clip number of each video is set to 128. Each word in the query sentence are embedded into a 300-dim vector by GloVe embedding~\cite{pennington2014glove}. The batch size is set to 100. For the 1D convolutional operation in {Feature Encoder}, the kernel size $K_{1D}$ and layer number $N_{conv}$ are set to 7 and 4, respectively. The number of {transformer block} is set to $N_{l}$ = 6. The hyper-parameter $\lambda, \beta$ and $\alpha$ in Eqs.~\ref{loss:reg}-\ref{loss:cls} are set to 0.5, 1.0 and 2.0, respectively. The parameter of the model is optimized with Adam optimizer with learning rate of $1\!\times\!10^{-4}$. We set the dropout ratio to $0.1$ for multi-head self-attention in the {video-language transformer}. To facilitate the implementation of the proposed model, the ground-truth values of starting and ending timestamps are normalized to [0, 1].

\subsection{Main Comparison}
To validate the effectiveness of the proposed \textbf{\emph{ViGT}}, we compare it with the state-of-the-art methods on the aforementioned three datasets, involving: \emph{\textbf 1) Proposal-based methods}: CTRL~\cite{gao2017tall}, MCN~\cite{anne2017localizing}, TGN~\cite{chen2018temporally}, CMIN~\cite{zhang2019cross}, SCDM~\cite{yuan2019semantic}, CBP~\cite{wang2020temporally}, 2D-TAN~\cite{zhang2020learning}, DRN~\cite{zeng2020dense}, MATN~\cite{zhang2021multi} CBLN~\cite{liu2021context}, MI-PUL~\cite{ding2022exploring}, MGPN~\cite{sun2022you}, HCLNet~\cite{zhang2022video}, and
\emph{\textbf 2) Proposal-free methods}: ABLR~\cite{yuan2019find}, TripNet~\cite{hahn2019tripping}, SM-RL~\cite{wang2019language}, DEBUG~\cite{lu2019debug}, TMLGA~\cite{rodriguez2020proposal}, LGI~\cite{mun2020local}, PMI~\cite{chen2020learning}, HVTG~\cite{chen2020hierarchical}, VSLNet~\cite{zhang2020vslnet}, BPNet~\cite{xiao2021boundary}, CPNet~\cite{li2021proposal}, DORi~\cite{rodriguez2021dori}, DRFT~\cite{chen2021end}, IVG~\cite{nan2021interventional}. 
It is well known that proposal-based methods achieve promising performance, but are accompanied by numerous proposal candidates and expensive post-processing. Proposal-free methods implemented in an end-to-end manner always retain a few hyperparameters, which can save computation cost.

\begin{table*}[t]
\centering
\footnotesize
\tabcolsep 21pt
\caption{Performance comparison on the YouCookII dataset.}
\begin{tabular}{c|c|c|cc|c}
	\toprule
	Method  &  Venue  & Core Tech& IoU@0.5$\uparrow$ & IoU@0.7$\uparrow$ & mIoU$\uparrow$ \\ \hline
	Random  & WACV'20 & --    &  1.72        &       0.60        &       -        \\
	TMLGA~\cite{rodriguez2020proposal}   & WACV'20 &  Attention    & 20.65       &       10.94       &     23.07      \\ \hline
	Ours   &    -    &Transformer&  \textbf{27.18}   &  \textbf{12.17}   & \textbf{27.61} \\ \bottomrule
\end{tabular}
\label{tab:youcook2_result}
\end{table*}

\begin{table*}[ht]
\centering
\footnotesize
\tabcolsep 19pt
\caption{Ablation studies of token, \textbf{FE} (Feature Encoder)} and \textbf{CMCA} (Cross-Modal Co-Attention) layers on the ANet-Captions dataset.
\begin{tabular}{ccc|c|c|c|c}
	\toprule
	\textbf{token} & \textbf{FE}  & \textbf{CMCA} & IoU@0.3$\uparrow$ & IoU@0.5$\uparrow$ & IoU@0.7$\uparrow$ & mIoU$\uparrow$ \\ \hline
	-       & $\checkmark$ & $\checkmark$  &       64.29       &       44.72       &       24.78       &     45.04      \\
	$\checkmark$  &      -       &       -       &       63.60       &       44.84       &       24.68       &     44.56      \\
	$\checkmark$  & $\checkmark$ &       -       &       64.32       &       46.05       &       25.67       &     45.23      \\
	$\checkmark$  &      -       & $\checkmark$  &       63.67       &       45.83       &       26.28       &     45.11      \\ \hline
	$\checkmark$  & $\checkmark$ & $\checkmark$  &  \textbf{64.59}   &  \textbf{46.71}   &  \textbf{26.90}   & \textbf{45.71} \\ 
	\bottomrule
\end{tabular}
\label{tab:abs_str}
\end{table*}

\begin{figure*}[t]
\centering
\includegraphics[width=1.0\linewidth]{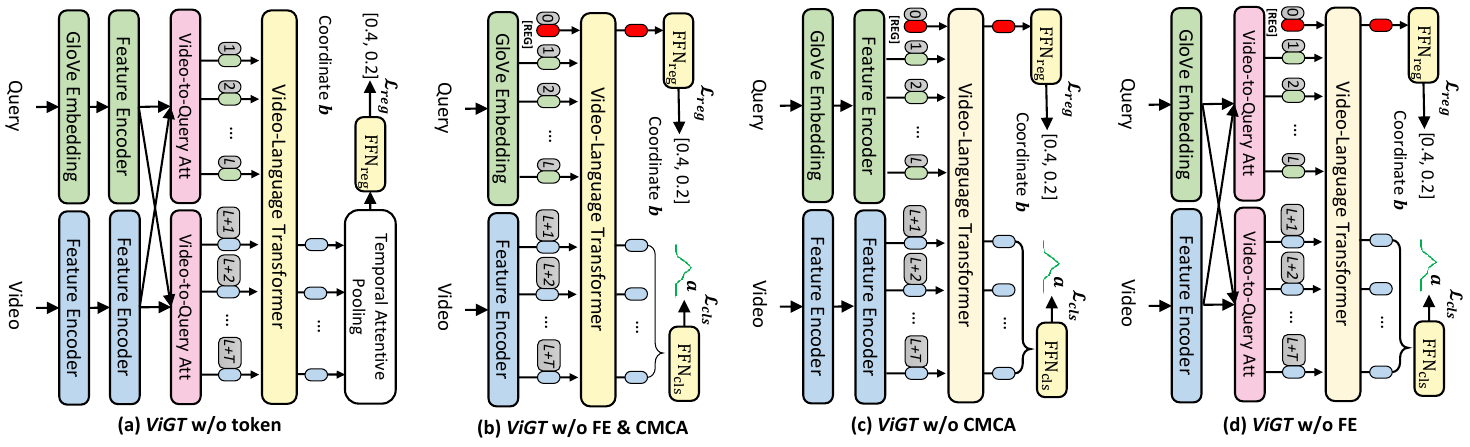}
\vspace{-1.0em}
\caption{Various architectures of ``\textit{\textbf{ViGT}}" in ablation experiments. ``\textit{\textbf{ViGT}} w/o token'' is used to verify the effectiveness of regression token [REG]. ``\textit{\textbf{ViGT}} w/o FE\&CMCA'' denotes the removal of both \textbf{FE} (Feature Encoder) and \textbf{CMCA} (Cross-Modal Co-Attention) layers in video-language encoder. ``\textit{\textbf{ViGT}} w/o CMCA'' and ``\textit{\textbf{ViGT}} w/o FE'' are used to verify the effects of \textbf{FE} and \textbf{CMCA} layers separately. 
}
\label{fig:abl}
\end{figure*}

Observing Tables~\ref{tab:anet_results}$\sim$\ref{tab:youcook2_result}, \textbf{\emph{ViGT}} achieves competitive performances. 
On the large-scale dataset ANet-Captions~\cite{krishna2017dense}, it performs the best mIoU with 45.71 on proposal-free methods including DRFT except DRFT$^{\dagger}$, where DRFT$^{\dagger}$ introduce extra data, \ie, depth and optical flow features. Even this, \textbf{\emph{ViGT}} surpasses DRFT$^{\dagger}$ at IoU@0.5. 
On the TACoS dataset, \textbf{\emph{ViGT}} achieves the best among proposal-free methods and outperforms most proposal-based methods except for MATN$^\ddagger$~\cite{zhang2021multi}, MGPN~\cite{sun2022you}, and HCLNet~\cite{zhang2022video}, where MATN$^\ddagger$ is a transformer-based model with MLM training technique (masked language modeling) to extract candidate proposals, MGPN utilizes more complicated coarse- and fine- grained features to generate candidate proposals, and HCLNet use hierarchical contrastive learning to align video and query. 
Our approach is a simple and light end-to-end transformer model without MLM training; it achieves comparable performance with MATN (\eg, 45.99 v.s. 45.64).
Turning to the challenging proposal-free comparison on the TACoS dataset again, \textbf{\emph{ViGT}} has more significant improvements than existing models including CPNet, \eg, lift 28.69 to 30.97 with 8.0\% improvement on mIoU, and lift 28.29 to 32.32 14.3\% improvement on IoU@0.5. 
On the YouCookII dataset, our proposed method achieves the best performance with ``IoU@0.7" of 12.17 and ``mIoU" of 27.61. 
These results demonstrate the effectiveness of our method for video grounding. Its superiority provides researchers with the token learning insight to regard regression models in this field. 
\begin{table}[tb]
\centering
\footnotesize
\tabcolsep 30pt
\caption{Ablation studies of the feature dimension $d$ in Video-Language Transformer on the ANet-Captions dataset.}
\begin{tabular}{c|ccc|c}
\toprule
\multicolumn{1}{c|}{Method} & IoU@0.3$\uparrow$ & IoU@0.5$\uparrow$ & IoU@0.7$\uparrow$ & mIoU$\uparrow$ \\ \hline
$d$=128           &       63.84       &       45.04       &       24.36       &     44.62      \\
$d$=256           &       64.50       &       45.44       &       25.84       &     45.20      \\
$d$=512           &  \textbf{64.59}   &  \textbf{46.71}   &  \textbf{26.90}   & \textbf{45.71} \\ \bottomrule
\end{tabular}
\end{table}

\begin{table}[t]
\centering
\footnotesize
\tabcolsep 30pt
\caption{Ablation studies of the layer number $N_{l}$ in Video-Language Transformer on the ANet-Captions dataset.}
\begin{tabular}{c|ccc|c}
\toprule
\multicolumn{1}{c|}{Method} & IoU@0.3$\uparrow$ & IoU@0.5$\uparrow$ & IoU@0.7$\uparrow$ & mIoU$\uparrow$ \\ \hline
$N_{l}$=1          &  \textbf{65.35}   &       44.87       &       22.44       &     44.62      \\
$N_{l}$=2          &       64.80       &       45.58       &       25.21       &     45.03      \\
$N_{l}$=4          &       64.32       &       46.08       &       25.77       &     45.27      \\
$N_{l}$=6          &       64.59       &  \textbf{46.71}   &  \textbf{26.90}   & \textbf{45.71} \\
$N_{l}$=8          &       63.58       &       45.53       &       25.76       &     44.92      \\ \bottomrule
\end{tabular}
\label{tab:abs_nvl}
\end{table}

\begin{wrapfigure}{l}{8.0cm}
\centering
\includegraphics[width=1.0\linewidth]{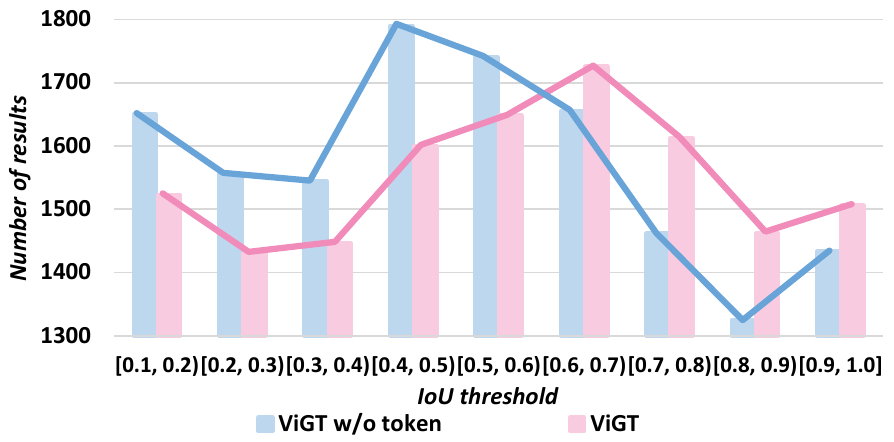}
\caption{The statistics of predicted results across different IoU ranges on the ANet-Captions dataset. Compared with ``\textbf{\emph{ViGT}} w/o token'', ``\textbf{\emph{ViGT}}'' predicts high-quality segments, especially at higher IoU ranges.}
\vspace{-1.0cm}
\label{fig:distri}
\end{wrapfigure}

\subsection{Ablation Study} 
In this subsection, we conduct adequate ablation studies to validate the effectiveness of \textbf{\emph{ViGT}}, involving: 1) learnable token [REG], 2) FE and CMCA layers in video-language encoder, 3) parameters $N_{l}$ and $d$ of video-language transformer, 4) parameter sharing in FE, 5) the order of FE and CMCA layers, 6) two objective losses $\mathcal L_{reg}$ and $\mathcal L_{cls}$, and 7) inference speed. The variants of the proposed model used in these ablation experiments are shown in Figure~\ref{fig:abl}.

\subsubsection{Learnable token} 
To verify the effectiveness of token [REG], we remove it and feed the transformed feature $\bm{\hat{V}^*}$$\in$$\mathbb{R}^{T\times d}$ 
into an attentive-regression module (widely used in recent proposal-free methods~\cite{mun2020local,li2021proposal}) for boundary prediction. 
This ablation model is named ``\textit{\textbf{ViGT}} w/o token'', and its architecture is shown in Figure~\ref{fig:abl} (a). 
Observing the first row of Table~\ref{tab:abs_str}, 
regression without token learning leads to obvious performance degradation at IoU@0.5/0.7. This meets our intention of using a token to learn more informative relations between  video and query sequences. 

In addition, we perform statistics on the prediction results across different IoU ranges. As shown in Figure~\ref{fig:distri}, we illustrate the IoU distribution of all the prediction results on the ANet-Captions dataset. 
From Figure~\ref{fig:distri}, we can obverse that ``\textbf{\emph{ViGT}}'' predicts more high-quality segments than ``\textbf{\emph{ViGT}} w/o token'', especially at the range of [0.8, 0.9). 
The higher the IoU, the more accurate the prediction result. 
The results reflect that the regression with token learning is more efficient than the multi-modal feature regression. 

\subsubsection{FE and CMCA layers} 
After that, we test the effect of feature encoder. 
In the setting of ``\textbf{\emph{ViGT}} w/o \textbf{FE\&CMCA}'', we input original video and query features into the Video-Language Transformer. We also test ``\textbf{\emph{ViGT}} w \textbf{FE}'' and ``\textbf{\emph{ViGT}} w \textbf{CMCA}''.
From Tables~\ref{tab:abs_str} and \ref{tab:anet_results}, we observe that ``\textbf{\emph{ViGT}} w/o \textbf{FE\&CMCA}'' surpasses most proposal-free methods. By adding \textbf{FE}, the ``\textbf{\emph{ViGT}} w \textbf{FE}'' shows a significant improvement on metrics IoU@0.3 and IoU@0.5, which means that the \textbf{FE} layer benefits the accuracy improvement of massive prediction results with low intersection values. The ``\textbf{\emph{ViGT}} w \textbf{CMCA}'' yields a significant boost on metric IoU@0.7 with high intersection values; it means that \textbf{CMCA} can enhance the correlation of crucial visual and textual cues. Anyway, the full model ``\textbf{\emph{ViGT}} w \textbf{FE\&CMCA}'' performs the best.

\subsubsection{Parameters of Video-Language Transformer} 
We test the layer number of transformer $N_{l}$ $\in$ $\{1,2,4,6,8\}$. As shown in Table~\ref{tab:abs_nvl}, the best performance is obtained with $N_{l}$ = $6$. 
We set it as the optimal setting. 
For the feature dimension $d$, as shown in Table~\ref{tab:abs_nvl}, the \textbf{\emph{ViGT}} gets the best results while $d=$512. 

\subsubsection{Parameter sharing in FE layer} 
Since the task targets to predict the coordinates of the queried action in the video, the model is required to understand the cross-modal correlation between video and query. In the feature encoding (FE) stage, the purpose of sharing weights is to map the features of video and query into the same feature space, and then facilitate the subsequent cross-modal correlation. 
If without the parameter sharing strategy, the feature encoding of each modality can be taken as independent feature embedding. Actually, the quality of original features is relatively good, and the independent feature refinement just has a slight effect on performance improvement. 
We conducted the experiment that the model without share weights strategy, and the results are listed in Table~\ref{tab:unshare}. We can see that sharing weights improves the performance on IoU=0.7 by a large margin (i.e., from 24.49 to 26.90). 
In other words, the parameter sharing strategy is used as a data preparation process for the subsequent cross-modal correlation. 
These results validate the effectiveness of parameter sharing in the feature encoder layer. 

\subsubsection{Order of FE and CMCA layers}
To verify the effectiveness of the order of FE and CMCA layers, we conducted the experiment of the model that first conducts CMCA and then implements FE. 
The experimental results are listed in Table~\ref{tab:abs_cmcafe}. 
We can see that if first conducts CMCA and then implements FE, the performance of the model will drop by a large margin (\eg, mIoU from 45.71 to 42.33, and IoU=0.7 from 26.90 to 20.96). 
As stated in the above discussion, the parameter sharing strategy benefits the subsequent cross-modal co-attention by first mapping them into a same feature space. 
Thus, we first use the FE to enhance the features of video and query through parameter sharing and then implement the CMCA. 

\begin{table}[t]
\centering
\footnotesize
\tabcolsep 19pt
\caption{Ablation studies of parameter sharing in \textbf{FE} (Feature Encoder) layer on the ANet-Captions dataset. ViGT-unshare represents that the weight of feature encoder is unshared.}
\resizebox{\linewidth}{!}{
	\begin{tabular}{c|c|c|c|c}
		\toprule
		Setting         &    IoU=0.3$\uparrow$ &     IoU=0.5$\uparrow$  &    IoU=0.7$\uparrow$ &   mIoU$\uparrow$  \\ \hline
		ViGT-unshare    &     64.57      &   44.96    &     24.49      &    45.17      \\
		\textbf{ViGT-share (Ours)} &  \textbf{64.59} &  \textbf{46.71} & \textbf{26.90} & \textbf{45.71} \\ \bottomrule
	\end{tabular}
}
\label{tab:unshare}
\end{table}

\begin{table*}[t!]
\centering
\footnotesize
\tabcolsep 19pt
\caption{Ablation studies of the order of \textbf{FE} (Feature Encoder) and \textbf{CMCA} (Cross-Modal Co-Attention) layers on the ANet-Captions dataset.}
\resizebox{1.0\linewidth}{!}{
	\begin{tabular}{c|c|c|c|c}
		\toprule
		Setting         &   IoU=0.3$\uparrow$ & IoU=0.5$\uparrow$  &    IoU=0.7$\uparrow$ &   mIoU$\uparrow$ \\ \hline
		ViGT (CMCA $\rightarrow$ FE)    &  61.13     & 40.83    &  20.96  & 42.33    \\
		\textbf{ViGT (FE $\rightarrow$ CMCA)} &  \textbf{64.59} & \textbf{46.71} & \textbf{26.90} &\textbf{45.71} \\ \bottomrule
	\end{tabular}
}
\label{tab:abs_cmcafe}
\end{table*}

\begin{table*}[t!]
\centering
\arrayrulecolor{black}
\footnotesize
\tabcolsep 20pt
\caption{Ablation studies of different losses on ANet-Captions dataset.}
	\begin{tabular}{ccc|ccc|c}
		\toprule
		$\mathcal{L}_{s\!-\!l1}$ & $\mathcal{L}_{giou}$ & $\mathcal{L}_{cls}$ & IoU@0.3$\uparrow$ & IoU@0.5$\uparrow$ & IoU@0.7$\uparrow$ & mIoU$\uparrow$ \\ \hline
		$\checkmark$       &          -           &          -          &       57.07       &       37.53       &       17.44       &     37.84      \\
		-             &     $\checkmark$     &          -          &       59.66       &       32.04       &       9.45        &     36.70      \\
		$\checkmark$       &          -           &    $\checkmark$     &       57.53       &       39.53       &       21.48       &     39.89      \\
		-             &     $\checkmark$     &    $\checkmark$     &  \textbf{65.89}   &       45.59       &       24.45       &     45.47      \\
		$\checkmark$       &     $\checkmark$     &          -          &       63.15       &  \textbf{47.47}   &       24.57       &     43.15      \\ \hline
		$\checkmark$       &     $\checkmark$     &    $\checkmark$     &       64.59       &       46.71       &  \textbf{26.90}   & \textbf{45.71} \\ \bottomrule
	\end{tabular}
	\vspace{-0.3cm}
\label{tab:abs_loss}
\end{table*}

\begin{table*}[t!]
\centering
\arrayrulecolor{black}
\footnotesize
\tabcolsep 20pt
\caption{Computation comparison of inference speed and required GPU memory on the ANet-Captions dataset.}
	\begin{tabular}{c|c|c|c|c|c}
		\toprule
		Model         & Batch size & Speed $\downarrow$  & Memory $\downarrow$ &    IoU@0.5$\uparrow$  &      mIoU$\uparrow$      \\ \hline
		2D-TAN~\cite{zhang2020learning}    & 100   &  0.0732s  &   8401M    &     44.51      &       -        \\
		\textbf{\textbf{\textit{ViGT}} (Ours)} & 100 & 0.0018s &   3755M   & \textbf{46.71} & \textbf{45.71} \\ \bottomrule
	\end{tabular}
\label{tab:gpu}
\end{table*}

\begin{figure*}[t]
\centering
\includegraphics[width=1.0\columnwidth]{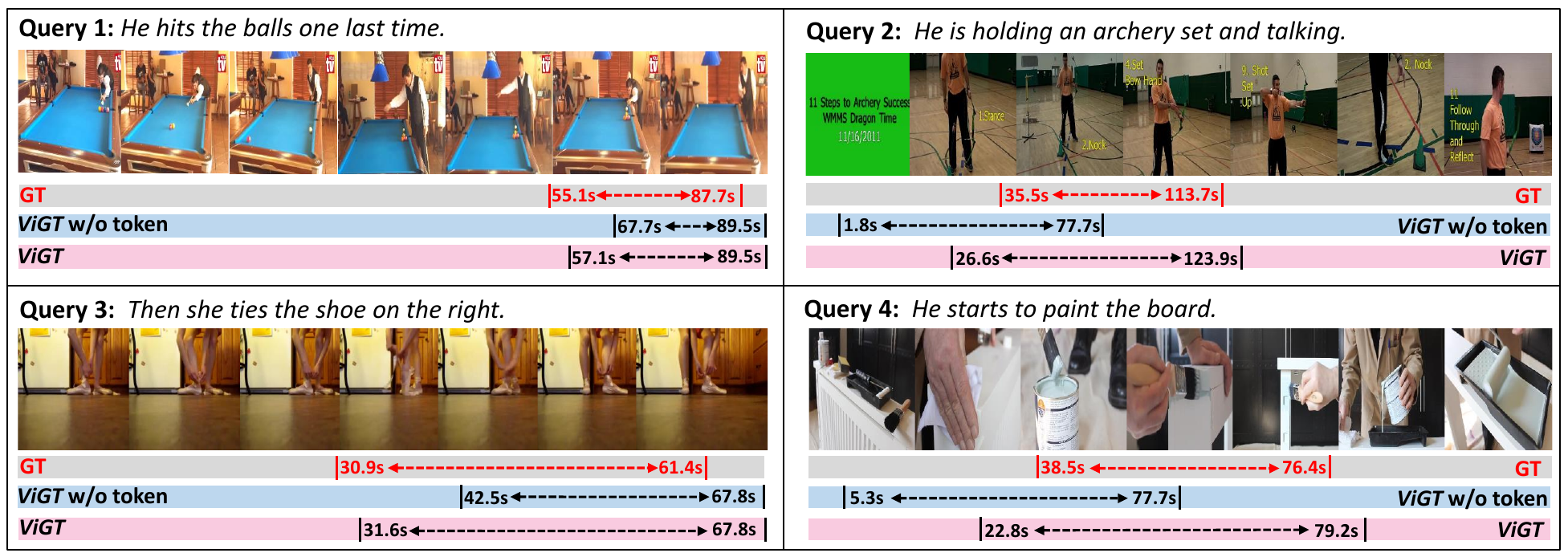}
\caption{Qualitative results of ``\textbf{\textit{ViGT}}'' and ``\textbf{\textit{ViGT}} w/o token'' on the ANet-Captions dataset. Compared with with ``\textbf{\textit{ViGT}} w/o token'', the proposed ``\textbf{\textit{ViGT}}'' locates the target segment more accurately in different scenarios.}
\label{fig:pred_vis}
\end{figure*}

\begin{figure*}[t!]
\centering
\includegraphics[width=1.0\linewidth]{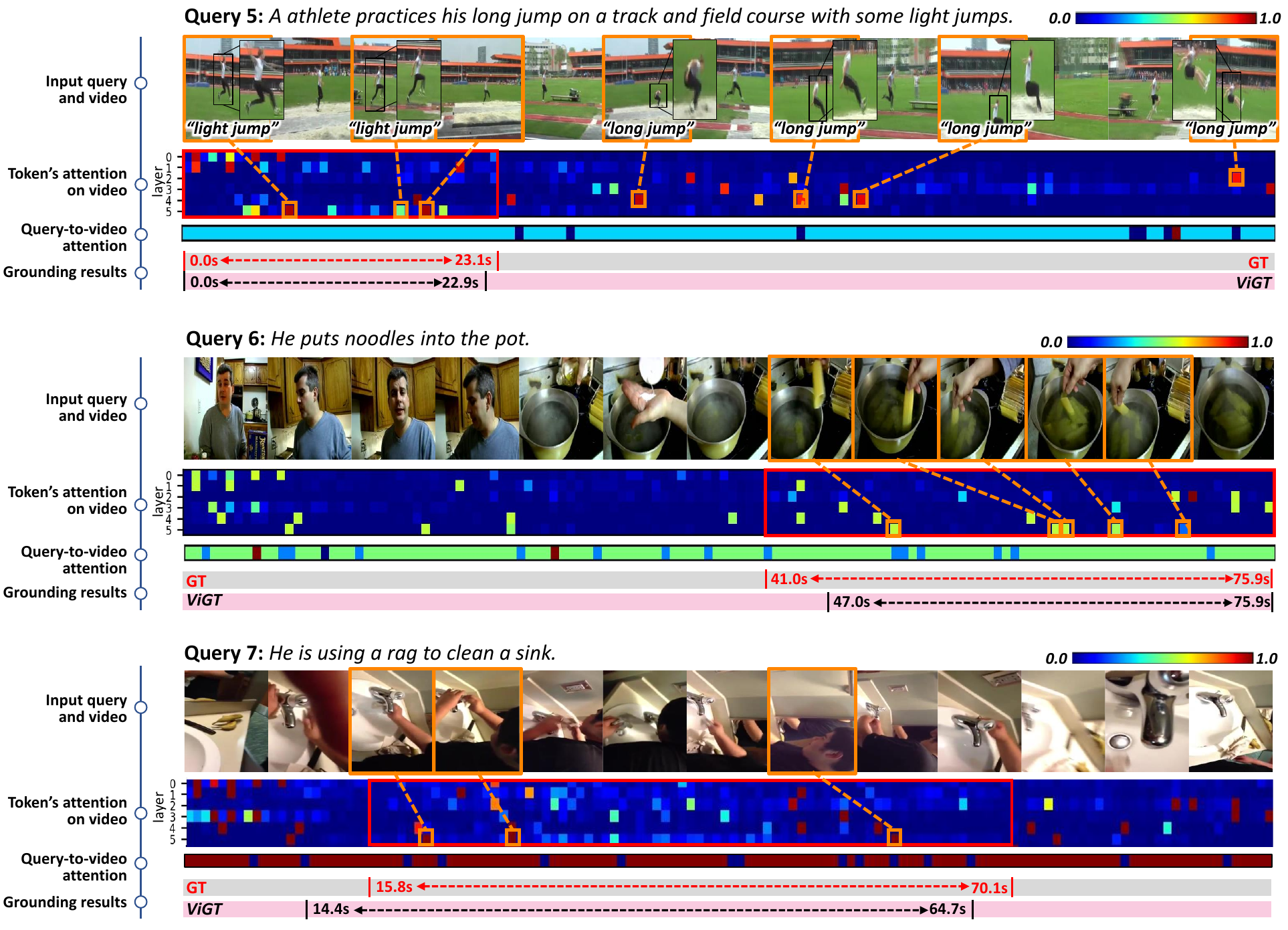}
\caption{Visualization of [REG] token's attention on the video in video-language transformer and query-to-video attention in {CMCA}. 
Different from \emph{query-to-video attention} of {CMCA} attending almost frames evenly, {the video-language transformer} performs different visual preference and progressively attends to more discriminative frames.}
\label{fig:vis_1}
\end{figure*}

\subsubsection{Objective losses $\mathcal L_{reg}$ and $\mathcal L_{cls}$} 
At first, we discuss the regression objective $\mathcal L_{reg}$, which consists of two terms $\mathcal{L}_{s\!-\!l1}$ and $\mathcal{L}_{giou}$. 
As shown in rows 1-2 of Table~\ref{tab:abs_loss}, the proposed model with either single $\mathcal{L}_{s\!-\!l1}$ or $\mathcal{L}_{giou}$ can achieve unsatisfactory performances. 
While injecting the classification loss $\mathcal{L}_{cls}$, under $\mathcal{L}_{s\!-\!l1}$\&$\mathcal{L}_{cls}$ and $\mathcal{L}_{giou}$\&$\mathcal{L}_{cls}$, the performances improve stably as shown in rows 3-4 of Table~\ref{tab:abs_loss}. Obviously, $\mathcal{L}_{cls}$ has a significant impact. Besides, please notice that $\mathcal{L}_{giou}$ is a new term originated from object detection. In the video grounding task, the model with only $\mathcal{L}_{giou}$ performs the worst, such as 9.45 at IoU@0.7, but the combinations of $\mathcal{L}_{giou}$ with any others all achieve promising performances, such as $\mathcal{L}_{s\!-\!l1}$\&$\mathcal{L}_{giou}$, $\mathcal{L}_{giou}$\&$\mathcal{L}_{cls}$, or $\mathcal{L}_{s\!-\!l1}$\&$\mathcal{L}_{giou}$\&$\mathcal{L}_{cls}$. $\mathcal{L}_{giou}$ is an effective auxiliary loss for this task. As a conclusion, \textbf{\emph{ViGT}} with the all the loss terms reaches a large margin improvement, \eg, raising the mIoU up to 45.71. 

\subsubsection{Inference speed of the proposed model} 
We compare our method with a typical proposal-based method 2D-TAN~\cite{zhang2020learning}.
In Table~\ref{tab:gpu}, the speed represents the average inference time per query, and the memory denotes the GPU memory consumption of each model. Compared with the 2D-TAN, the \textit{\textbf{ViGT}} consumes less GPU memory, \ie, 3755M \vs 8401M. 
In addition, the proposed \textit{\textbf{ViGT}} outperforms 2D-TAN by a large margin in the term of inference speed, \ie, 0.0018s \vs 0.0732s per query. The faster inference speed and lower memory consumption demonstrate the outstanding superiority of \textit{\textbf{ViGT}} without extra proposal post-processing. 

\subsection{Qualitative Results} 
\begin{figure*}[t!]
\centering
\includegraphics[width=1.0\linewidth]{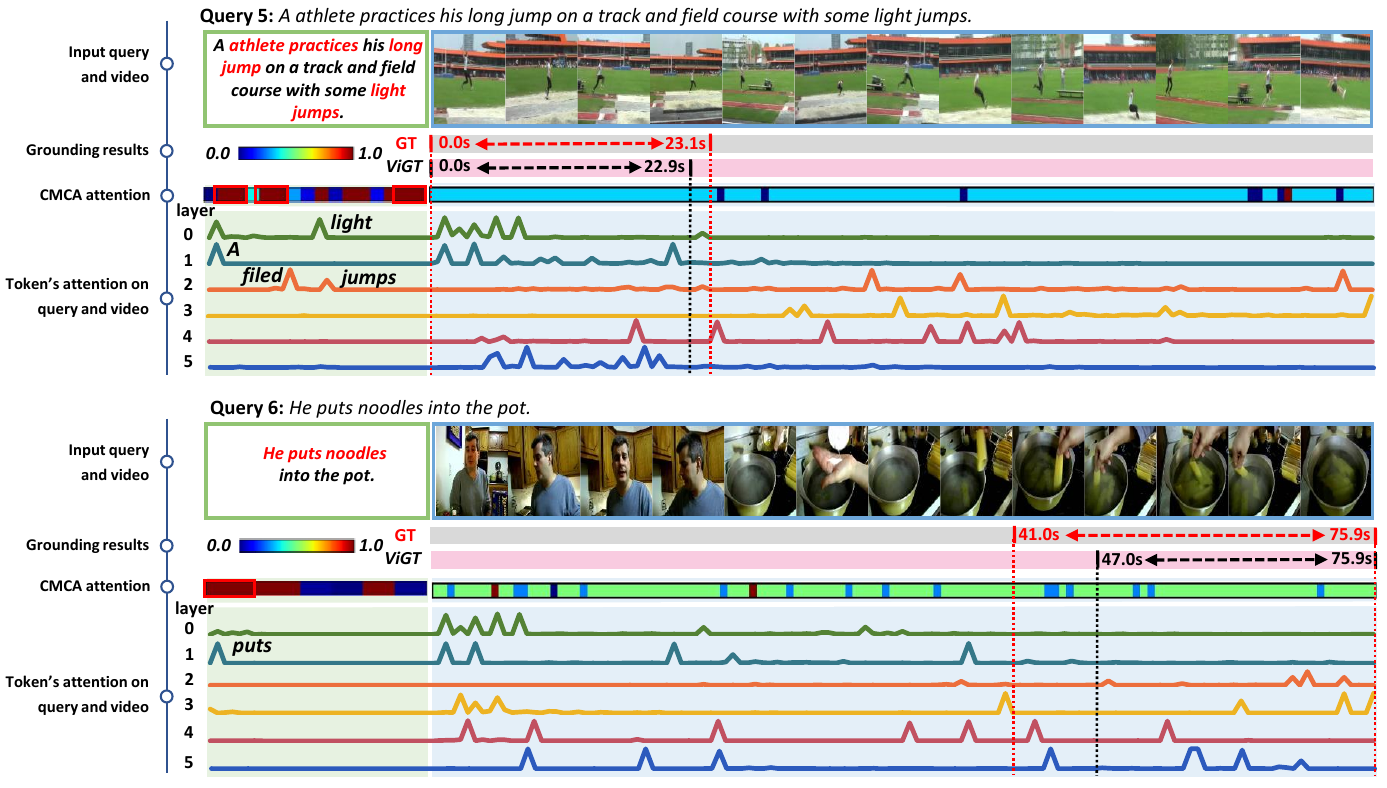}
\vspace{-0.3cm}
\caption{Visualization of [REG] token's attention on both video and query in video-language transformer and {CMCA} modules.
In this figure, we display the attention distribution in the term of attention curves. For \emph{Queries} 5 and 6, the {CMCA} and {video-language transformer} exhibit opposite properties, where the {CMCA} benefits for discovering query cues and the {video-language transformer} benefits for discovering video cues. In summary, \textbf{\emph{ViGT}} progressively searches relevant visual and textual cues, and predicts the target moments accurately. 
}
\label{fig:vis_2}
\end{figure*}

To demonstrate the effectiveness of interpretability of \textbf{\emph{ViGT}}, we display some video samples in Figures~\ref{fig:pred_vis}, \ref{fig:vis_1}, and \ref{fig:vis_2}.
At first, as shown in Figure~\ref{fig:pred_vis}, we display four qualitative results from the ANet-Captions dataset. For ``$Query$ 1: \textit{He hits the balls one last time.}'' and ``$Query$ 3: \textit{Then she ties the shoe on the right.}'', both ``\textbf{\emph{ViGT}} w/o token'' and \textbf{\emph{ViGT}} locate the target moment, but \textbf{\emph{ViGT}} predicts more accurate boundaries than ``\textbf{\emph{ViGT}} w/o token''.
Compared with the ground-truth and \textbf{\emph{ViGT}}, ``\textbf{\emph{ViGT}} w/o token'' predicts much more narrowing boundaries in $Queries$ 1\&3. 
For ``$Query$ 2: \textit{He is holding an archery set and talking.}'' and ``$Query$ 4: \textit{He starts to paint the board.}'', obviously, the ``\textbf{\emph{ViGT}} w/o token'' predicts the inexact moments, while \textbf{\emph{ViGT}} precisely locate the action in the videos. 
From the above qualitative results, the token can effectively aggregate global semantics of video and query for video grounding. 

In Figure~\ref{fig:vis_1}, we illustrate the [REG] token's attention on the visual sequence step by step. The upper six rows display the token's attention maps on the video from the layer-0 to layer-5 of the video-language transformer, and the bottom row reflects the \emph{query-to-video} attention map of {CMCA}. 
Taking $Queries$ 5$\sim$7 as examples, either for the videos in $Queries$ 5$\sim$6 having continuous changing frames or the video in $Query$ 7 having similar frames, the \emph{query-to-video} pays attention to almost frames evenly. This is not in line with the queried moment. By comparison, the token's attention shows significant disparity, which searches the potential segments in the six layers of transformer progressively and finally predicts the target segment accurately.

To further display the token's effectiveness, we show the textual attention curves of {CMCA} and {video-language transformer} in Figure~\ref{fig:vis_2}. 
Observing the {CMCA} and the video-language transformer, the contributive textual attention is conducted by {CMCA}. 
For the {video-language transformer}, it no longer pays attention to any word after the 3-$th$ layer of \emph{Query} 5 and the 2-$th$ layer of \emph{Query} 6. 
The {CMCA} and {video-language transformer} perform oppositely, where the {CMCA} module benefits for discovering query cues and the {video-language transformer} modules benefits for discovering video cues. 
To further validate this, we also show the visual attention curves on video in Figure~\ref{fig:vis_2}. 
Taking $Query$ 5 as an example, it is a challenging video, in which visual appearances of ``\emph{light jump}'' and ``\emph{long jump}'' are easily confusing. 
The textual attention of {CMCA} is useful, but the visual attention of {CMCA} is useless as it attends almost frames eventually. By comparison with the {CMCA}, the visual attention of transformer is much more distinctive and useful. Specifically, the {video-language transformer} firstly attends to the segments covering ``\emph{slight jump}'' at $\{0, 1\}$-$th$ layers. Then, it attends to the segments covering ``\emph{long jump}'' at $\{2, 3, 4\}$-$th$ layers. 
Eventually, \textbf{\emph{ViGT}} locates the correct segment of ``\emph{slight jumps}'' at the last layer.

\section{Conclusion}\label{sec:conclusion}
In this paper, we investigate a learnable regression token for video grounding and propose a novel transformer-based proposal-free framework named \emph{\textbf{ViGT}}. Token learning is a new methodological paradigm for video grounding that predicts the target moment using a learnable token rather than multi-modal features or cross-modal features, as in previous works. 
Extensive experiments demonstrate that the token can learn informative semantics from videos and queries.  
The visualization results also show the interpretability of our method. 

\Acknowledgements{This work was supported by the National Natural Science Foundation of China (72188101, 62020106007, and 62272144, and U20A20183), and the Major Project of Anhui Province (202203a05020011).}



{
\bibliography{abrv.bib}
\bibliographystyle{scis}
}



\end{document}